\documentclass[conference]{IEEEtran}
\IEEEoverridecommandlockouts
\usepackage{cite}
\usepackage{amsmath,amssymb,amsfonts}
\usepackage{algorithmic}
\usepackage{graphicx}
\usepackage{textcomp}
\usepackage{xcolor}
\usepackage{caption}
\usepackage{subfigure}
\usepackage{graphicx}
\usepackage{algorithm}
\usepackage{booktabs} 
\usepackage{multirow}
\usepackage{threeparttable}
\usepackage{placeins}
\usepackage{stfloats}
\usepackage{float}

\def\BibTeX{{\rm B\kern-.05em{\sc i\kern-.025em b}\kern-.08em
    T\kern-.1667em\lower.7ex\hbox{E}\kern-.125emX}}
\begin{document}

\title{Operation-level Progressive Differentiable Architecture Search}

%
%
%

\author{
	\IEEEauthorblockN{ Xunyu Zhu$^{1,2}$, Jian Li$^{1*}$ \thanks{$^*$Corresponding author.}, Yong Liu$^3$,  Weiping Wang$^{1,2}$}
	\IEEEauthorblockA{$^1$ Institute of Information Engineering, Chinese Academy of Sciences, Beijing, China}
	\IEEEauthorblockA{$^2$ School of Cyber Security, University of Chinese Academy of Sciences, Beijing, China }
	\IEEEauthorblockA{$^3$ Gaoling School of Artificial Intelligence, Renmin University of China, Beijing, China}
	}


\maketitle

\begin{abstract}

Differentiable Neural Architecture Search (DARTS) is becoming more and more popular among Neural Architecture Search (NAS) methods because of its high search efficiency and low compute cost. However,  the stability of DARTS  is very inferior, especially skip connections aggregation that leads to performance collapse. Though existing methods leverage Hessian eigenvalues to alleviate skip connections aggregation, they make DARTS unable to explore architectures with better performance.  In the paper, we propose operation-level progressive differentiable neural architecture search (OPP-DARTS) to avoid skip connections aggregation and explore better architectures simultaneously.  We first divide the search process into several stages during the search phase and increase candidate operations into the search space progressively at the beginning of each stage. It can effectively alleviate the unfair competition between operations during the search phase of DARTS by offsetting the inherent unfair advantage of the skip connection over other operations. Besides, to keep the competition between operations relatively fair and select the operation from the candidate operations set that makes training loss of the supernet largest.  The experiment results indicate that our method is effective and efficient. Our method's performance on CIFAR-10 is superior to the architecture found by standard DARTS, and the transferability of our method also surpasses standard DARTS.  We further demonstrate the robustness of our method on three simple search spaces, i.e., \textbf{S2, S3, S4}, and the results show us that our method is more robust than standard DARTS. Our code is available at https://github.com/zxunyu/OPP-DARTS.

\end{abstract}


\section{Introduction}

The development of machine learning \cite{DBLP:conf/aaai/LiLW20,DBLP:conf/ijcai/LiLY019,DBLP:conf/icml/0001WM21} and deep learning \cite{726791,NIPS2012_c399862d,simonyan2015deep} has promoted the revolution in the field of image classification. However, the cost of neural network architectures designed by hand is very high because it needs experienced experts to design high-performance network architectures, which  means that network architecture design is unable to common for people. The  problem impedes the development of deep learning. Fortunately, the emergence of neural architecture search (NAS) effectively alleviates the problem. It  can make neural network design automatically  and design high-quality networks that can be par with manually designed architectures.

Architecture search methods based on reinforcement learning and evolutionary algorithms are the first architecture search methods to be proposed.  \cite{45826} proposes to leverage reinforcement learning to sample subnetworks and make the validation loss as the reward to help NAS select the best sub-architecture.
\cite{real2019regularized} proposes to leverage evolutionary algorithms to explore subnetworks with better performance. However, these methods based on reinforcement learning and evolutionary algorithms cost are cumbersome. NASNet \cite{2017arXiv170707012Z} proposes that NAS can select cells instead of the entire networks and stack cells to build networks. The method reduces computational consumption by compressing search space.


\begin{figure*}
	\centering
	\includegraphics[height=5cm,width=\linewidth]{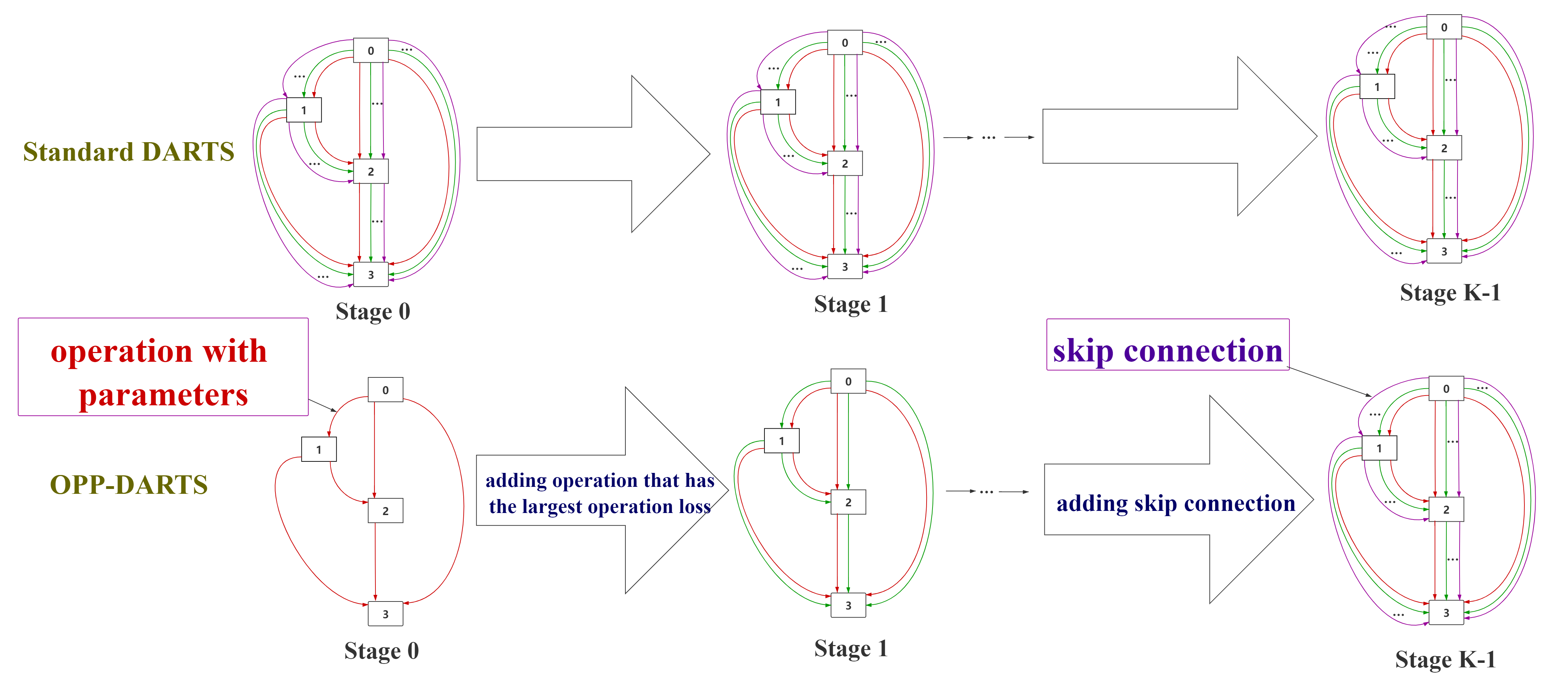}
	\caption{The bottom figure is an overview of OPP-DARTS and the above figure shows the search process of standard DARTS. }
	\label{opp-darts}
\end{figure*}

This paper proposes operation-level progressive differentiable architecture search (OPP-DARTS) to alleviate skip connections aggregation. Figure \ref{opp-darts} shows the basic procedure of OPP-DARTS. We first divide the search phase into several stages. In the first stage, we select one candidate operation with parameters to join in the supernet. After the first operation joined in the supernet, we train the supernet several epochs until the next stage. In the next stage, we select another operation to join the supernet and train the supernet. We repeat this action at every stage. Then we train the supernet until the final epoch. In addition, different operations are unfair to compete with each other during the training phase, such as $3 \times 3\_sep\_conv$ and $skip\_connect$. DARTS always would like to select $skip\_connect$  rather than $3 \times 3\_sep\_conv$, which means when $skip\_connect$ joins the supernet earlier than other operations, the supernet prefers to select $skip\_connect$. Therefore, we make the operation loss as the criteria to help the supernet select corresponding operation, i.e., the candidate operation that makes operation loss largest will be selected to join in the network. 

OPP-DARTS has demonstrated its effectiveness and efficiency in Experiments (Section \ref{exp}). We firstly search architecture on CIFAR-10 by using similar training settings with DARTS.  Then, we evaluate the searched architecture on CIFAR-10 and ImageNet, and the training settings are also the same with DARTS. When evaluating it on CIFAR-10, the test error of the searched architecture is $2.63\%$ with 3.8M parameters, and the performance is much better than standard DARTS. To verify the transferability of OPP-DARTS, we transfer the searched architecture to train on ImageNet, and the performance achieved on ImageNet shows that our method outperforms standard DARTS. At last, we further demonstrate the robustness of our method. We ran our method on three simple search spaces, i.e., \textbf{S2}, \textbf{S3}, \textbf{S4}, and the final results indicate that the robustness of our method is also better than standard DARTS.

%
%
%

\section{Related Work}

DARTS is a great innovation that relaxes neural architecture search into a continuous problem, i.e., makes NAS differentiable. Furthermore, this dramatically  reduces computation  cost and can search architecture with good performance in a single GPU day. However, DARTS has an unavoidable problem, and the problem prevents DARTS from being applied. The problem is skip connections aggregation, i.e., the architectures searched by DARTS are filled with excessive skip connections, and this phenomenon makes DARTS performance collapse. Some previous works have dealt with the problem. \cite{Zela2020Understanding}  proposes Hessian eigenvalues as a signal  to monitor skip connections aggregation. SDARTS\cite{pmlr-v119-chen20f} regularizes Hessian eigenvalues by increasing perturbations  to alleviate skip connections aggregation. P-DARTS \cite{Chen_2019_ICCV} draws up rules manually to  alleviate skip connections aggregation.   Existing methods regard  Hessian eigenvalues or the number of skip connections as the signal to guide to alleviate skip connections aggregation. These methods make some rules by hand or regularize Hessian eigenvalues to keep the stability of DARTS.    However, these methods seek quick success and instant benefits because they make DARTS unable to explore architectures with better performance.  DARTS- \cite{chu2021darts} adds an auxiliary skip connection to alleviate skip connections aggregation. The method is simple, whereas memory consumption will increase because of the extra skip connection.


\section{Method}

\subsection{Preliminary of DARTS}

Cell-based NAS methods \cite{2017arXiv170707012Z,2017arXiv171200559L,2018arXiv180201548R} have been proposed to learn cells instead of architectures to reduce computation overhead in the process of architecture search. The networks consist of many  same cell structures. A cell is similar to a directed acyclic graph (DAG), and it contains $N$ nodes, i.e., two input nodes, some intermediate nodes, and a single output node. Each node is a  latent representation symbolized as $x^{(i)}$, and each directed edge denoted as $(i,j)$ represents an  information flow that an operation $o^{(i,j)}$ in the directed edge $(i,j)$ transfers information from node  $x^{(i)}$ to node  $x^{(j)}$. Differentiable Architecture Search (DARTS) \cite{2018arXiv180609055L}  relaxes the selection problem of operations as a continuous optimization problem. There is a candidate operations space $\mathcal{O} $, where $o \in \mathcal{O} $ represents a candidate  operation, e.g., $skip\_connect, 3 \times 3\_sep\_conv$, and so on. In DARTS \cite{2018arXiv180609055L}, each edge consists of a set of operations from $\mathcal{O} $, and these operations are weighted by architecture parameters $\alpha^{i,j}$, and it can be formulated as:
\begin{equation}
	\label{fm}
	\bar{o}^{(i, j)}(x)=\sum_{o \in \mathcal{O}} \frac{\exp \left(\alpha_{o}^{(i, j)}\right)}{\sum_{o^{\prime} \in \mathcal{O}} \exp \left(\alpha_{o^{\prime}}^{(i, j)}\right)} o(x),
\end{equation}
where $0 \le i < j \le N-1$. The input nodes of the cell take the output  from the previous two cells as input, and the output node contacts all intermediate nodes as the output of the cell. Each intermediate node is be obtained by its predecessors, i.e. $x^{(j)}=\sum_{i<j} o^{(i, j)}\left(x^{(i)}\right)$. Finally, the architecture search of DARTS becomes a bi-level optimization problem:
\begin{align*}
	& \min _{\alpha} \quad \mathcal{L}_{v a l}\left(w^{*}(\alpha), \alpha\right) \\	
	& \text { s.t. } \quad w^{*}(\alpha)=\operatorname{argmin}_{w} \mathcal{L}_{\text {train }}(w, \alpha),
\end{align*}
where $\mathcal{L}_{v a l}$ and $\mathcal{L}_{\text {train }}$ are validation and training loss, $w$ is the network weights, and $\alpha$ is the architecture weights. According to DARTS\cite{2018arXiv180609055L}, the bi-level optimization problem is solved by a first/second-order approximation.  When the search process is nearly finished, an  optimal substructure is obtained based on the architecture weights $\alpha$, i.e.,  $o^{(i, j)}=\operatorname{argmax}_{o \in \mathcal{O}} \alpha_{o}^{(i, j)}$.

\subsection{Increasing Operations  Progressively}


To eliminate unfairness between operations, we propose operation-level progressive differentiable architecture search, briefly called "OPP-DARTS". In other words, we increase operations progressively to expand the search space during the search phase of DARTS, as illustrated in Figure \ref{opp-darts}, and the OPP-DARTS is detailed in Alg. \ref{alg1}. Firstly, the search phase of DARTS is divided into several stages denoted as $stage \ 1...K$, each stage consists of $T$ epochs, and one operation will be increased into the search space of DARTS  every stage.

Then, we will begin increasing operations to enlarge search space. Because the supernet must own an operation with parameters to make itself trainable, we must select an operation with parameters at  $stage \ 0$. We first define candidate operations with parameters set as $\mathcal{O}_{pm}$. Furthermore, we select an operation $o$ from  candidate operations set $\mathcal{O}_{pm}$. Then, we increase it into the supernet and begin to train the supernet until the next stage.

Afterwards, we will define the  candidate operations set as $\mathcal{O}$, and the candidate operations set $\mathcal{O}$ includes all candidate operations except the operation  selected in $stage \ 0$ and $skip\_connect$. At the beginning of each stage, we select an operation from candidate operations  set $\mathcal{O}$, and increase it into the supernet to train, i.e.,
\begin{align}
	\min _{\alpha} \quad & \mathcal{L}_{v a l}\left(o, \Omega , w^*, \alpha\right) \\	
	\text { s.t. } \quad & w^* =\operatorname{argmin}_{w} \mathcal{L}_{\text {train }}(o,\Omega , w, \alpha) \\
	& o \in \mathcal{O},
\end{align}
where $o$ is the operation selected at the stage and $\Omega$ is the search space of the supernet. In this way, these operations increased into the supernet early have the advantage compared to those increased into the supernet lately because these operations that are increased into the supernet early can learn more valuable knowledge than those operations. We can  alleviate the unfair natural  advantages between operations to make operations compete fairly. When we pick operations at random from the candidate operations set $\mathcal{O}$, it may cause a negative effect if we select operations with natural advantages compared with the others to increase it into the supernet earlier than other operations. Thus, we need to design a criterion to guide us to select operations, and Section \ref{opl} shows a criterion to alleviate the problem.

In  $stage \ K-1$, we increase $skip\_connect$ into the supernet because skip connection has the most significant  natural advantages than other operations. By increasing it into the supernet, at last, we can make other operations offset the natural advantages of skip connection to make operations compete fairly.

\subsection{Operation Loss}
\label{opl}

By increasing operations progressively, we can make operations compete fairly during the search phase. However, the problem still  needs to be solved, i.e., when we select operations from the candidate operations set $\mathcal{O}$ at each stage, we need to decide to select which operation to increase in the supernet.  The problem is very critical  because if we select operations with natural advantages compared with the others to increase it into the supernet earlier than other operations, it will negatively impact it. Thus, we need to design an applicable criterion to guide us to make operation selection.

\cite{wang2021rethinking} indicates that edges share the same optimal feature map in a cell, and it  means   the edge feature graphs will be closer and closer as the network converges. The feature map of an edge can be represented as Eq. \ref{fm}. At initialization, if the feature map on an operation is close to the optimal feature map, the architecture parameter $\alpha$ of the operation will become larger at the beginning of the architecture search. It will result in unfairness in initialization.  To keep operations competing  fairly, we select the operation which can make the supernet gain the largest training loss after the operation is increased  into the supernet at the beginning of each stage, called "operation loss", i.e.,
\begin{equation}
	\label{opls}
	o = argmax_ {o \in \mathcal{O}} \mathcal{L} _{train}(o,w,\alpha ),
\end{equation} 
where $o$ is the operation that is selected at the stage. When we use "operation loss" as a criterion to guide us to select an operation at the beginning of each stage, we can make feature maps of operations relatively close to the optimal feature map before they begin to compete with each other. We increase the operation that owns the largest "operation loss" to train at each stage so that the operation can be more closed to the optimal feature map. Thus, it can alleviate unfairness in initialization.  Furthermore, the search process of DARTS can be  formalized as follows:
\begin{align}
	\min _{\alpha} \quad & \mathcal{L}_{v a l}\left(o, \Omega , w^*, \alpha\right) \\	
	\text { s.t. } \quad & w^* =\operatorname{argmin}_{w} \mathcal{L}_{\text {train }}(o,\Omega , w, \alpha) \\
	& o = argmax_ {o \in \mathcal{O}} \mathcal{L} _{train}(o,\Omega,w,\alpha ),
\end{align}
where $\Omega$ is the search space of the supernet. In addition, the feature map of operation is saved in the parameters of operation. Training  loss is used to optimize network parameters, i.e., training  loss has a maximum  correlation with the  feature map of operation, so we select training loss instead of validation loss.

 We know that skip connection owns a natural advantage because it can make the network coverage faster, and it  means that skip connection owns a  more significant advantage than other operations, so we increased it into the supernet at the final stage to keep operations compete fairly.

\begin{algorithm}
	\caption{Operation-level Progressive Differentiable Architecture Search (OPP-DARTS)}
	\label{alg1}
	\begin{algorithmic}
		\REQUIRE     The  search space of the supernet $\Omega$,  all candidate   operations set $\Re  $.
		
		
		\STATE Select  $\vartheta  \in \mathcal{O}_{pm}$ based on Eq. (\ref{opls}) to join in $\Omega$;
		\STATE Initialize $\mathcal{O} = \Re  \setminus  \{skip\_connect,  \vartheta \}$
		
		
%
%
		\FOR {$i=1 \ to \ E$}{
			\IF {$ i \ \% \ T == 0 $ and $ i \le  T \times k$   }{
				\STATE Select  $o \in \mathcal{O}$ based on Eq. (\ref{opls})  to join in $\Omega$;
				\STATE Remove $o$ from $O$;
			}
			\ENDIF
			
			\IF {$ i == T \times (k+1) $   }{
				\STATE Select  $skip\_connect $   to join in $M$;
			}
			\ENDIF

			\STATE Update  architecture parameters $\alpha$ by descending $\bigtriangledown _{\alpha } \mathcal{L} _{val} (w,\alpha )$;
			
			\STATE Update weights $w$  by descending $\bigtriangledown _{w } \mathcal{L} _{train} (w,\alpha )$;
		}
		
		\ENDFOR

	\end{algorithmic}
\end{algorithm}


\subsection{Discussion}

In the paper, our method is proposed to deal with performance collapse in DARTS arising from skip connections aggregation. Our method owns two contributions mainly, i.e., operation-level increasing operations progressively and operation loss. The first contribution is a very significant  innovation to alleviate skip connections aggregation. Compared with other works to alleviate performance collapse in DARTS  by indicator-based methods (e.g.,  R-DARTS \cite{Zela2020Understanding}, SDARTS \cite{pmlr-v119-chen20f}), we alleviate performance collapse by optimizing the search space of DARTS to keep operations compete with each other fairly. Thus the view of our research is very novel. The second contribution is proposed to solve the operation selection problem at the beginning of each stage.  If we select operations from candidate  operations set at random, operations will suffer  from unfairness in initiation, making  DARTS work worse. By operation loss, we can alleviate unfairness in the initiation and make DARTS work well. We obtain a good performance (a test error of $2.63\%$) on CIFAR-10 by combining two contributions, and the result indicates that our method makes a remarkable improvement in accuracy on standard DARTS. At the same time,  the transferability and robustness of our method are also demonstrated to be better than standard DARTS.



\section{Experiments}
\label{exp}

We first search normal and reduction cells on CIFAR-10 \cite{krizhevsky2009learning} in Section \ref{as_10} . Then, we stack the selected cells to build a new deeper network to evaluate its performance on CIFAR-10 in Section \ref{ae_10}. Further, Section \ref{ae_in} shows the transferability of OPP-DARTS by stacking cells to build an even deeper network and then evaluating it on ImageNet. Finally, we verify the robustness of our methods by searching architectures on some simple search space, i.e., \textbf{S2, S3, S4} in Section \ref{rbt}. 

\begin{figure*}[]
	\centering
	\begin{tabular}{@{}c@{\hspace{1mm}}c@{\hspace{1mm}}c@{}}
		
		\includegraphics[trim=10mm 13mm 0mm 13mm, clip, width=7cm]{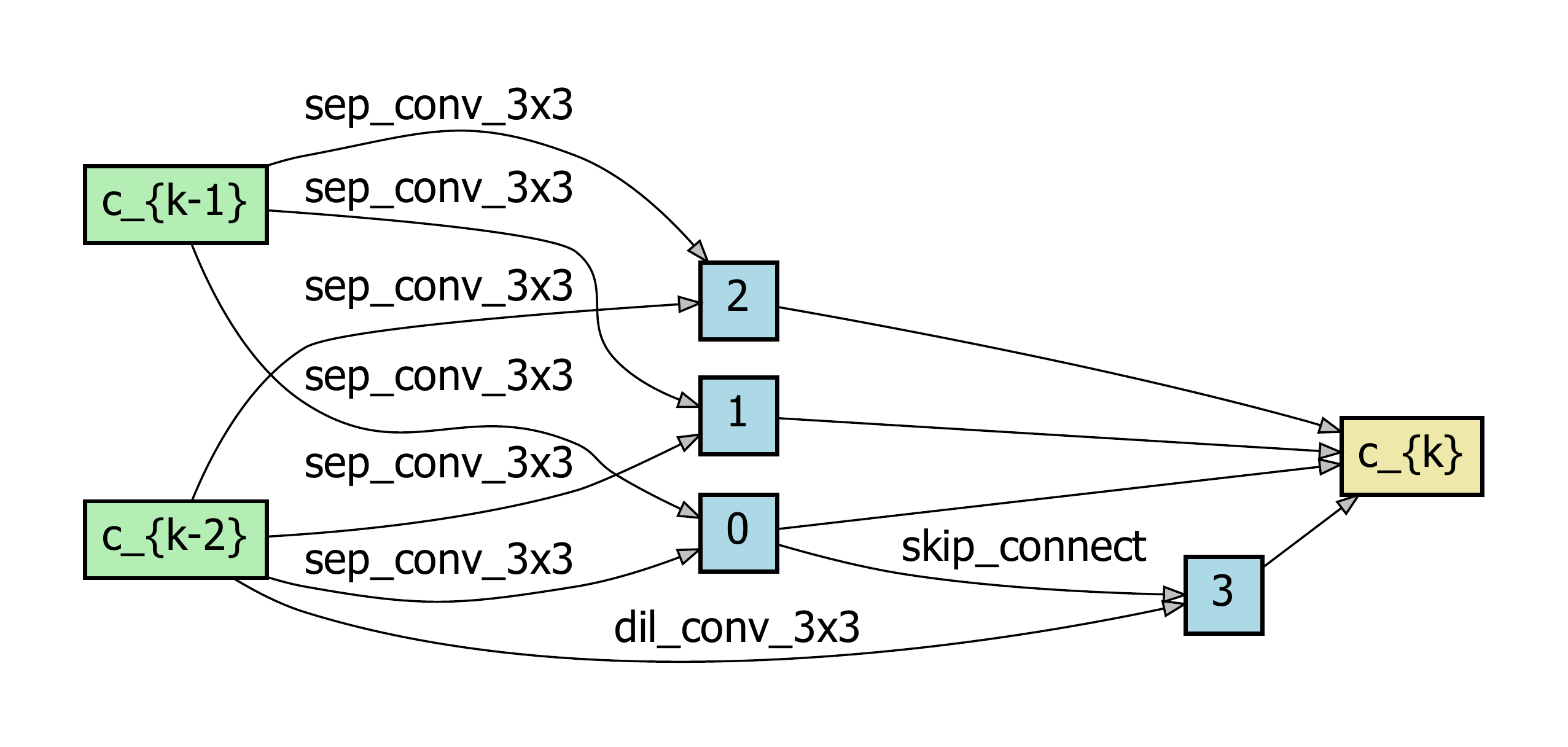} &
		\includegraphics[trim=0mm 13mm 10mm 13mm, clip,  width=7cm]{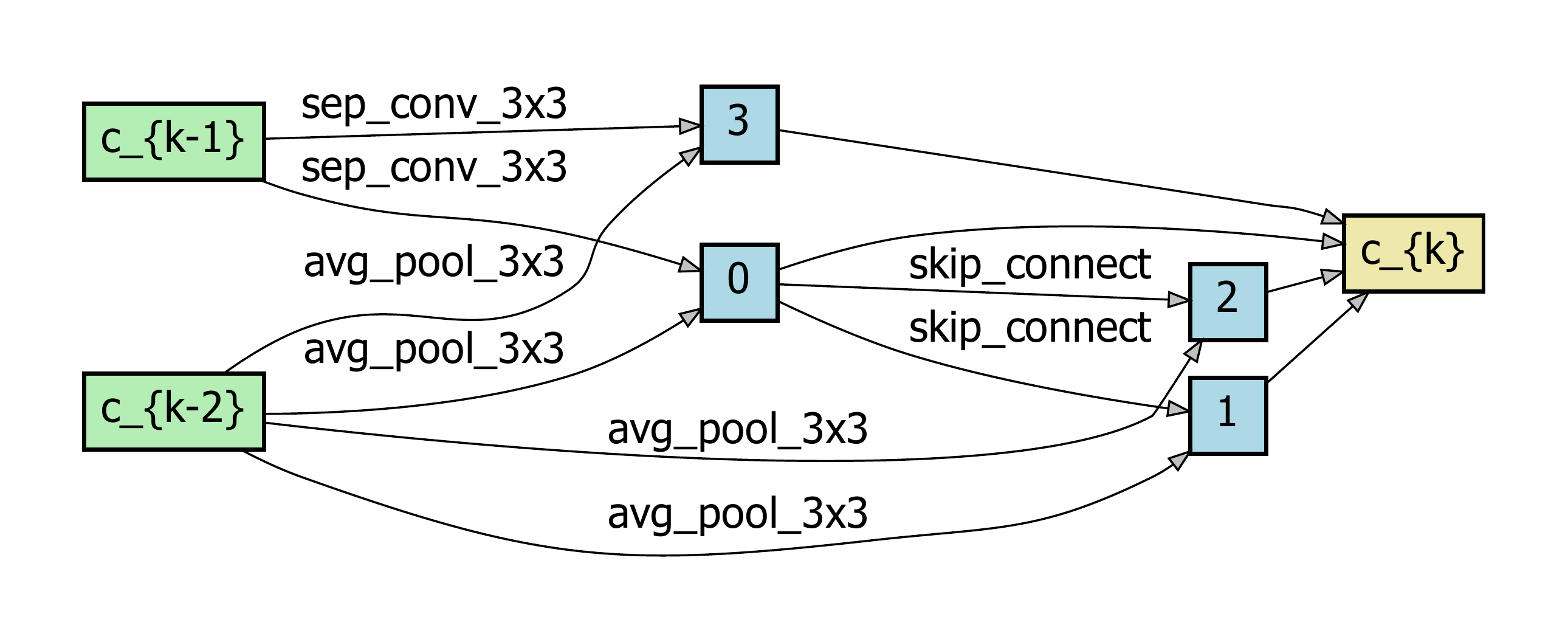} \\
		
		\small (a) Normal cell &
		\small (b) Reduction cell \\
		
	\end{tabular}
	\caption{Best cells are found by OPP-DARTS on CIFAR-10 (a). Normal cell is found by OPP-DARTS. (b). Reduction cell is found by OPP-DARTS.  }
	\label{fig_4}
	\vspace{-10pt}
\end{figure*}

\subsection{Architecture Search on CIFAR-10}
\label{as_10}

CIFAR-10  is an images dataset that contains $50K$ training images and $10K$ test images.  In the next moment, we will introduce our search space. Our search space consists of eight operation, i.e.,  $max\_pool\_3 \times 3, avg\_pool\_3 \times 3, zero, sep\_conv\_3 \times 3, sep\_conv\_5 \times 5, dil\_conv\_3 \times 3, dil\_conv\_5 \times 5$,  it is the same with DARTS.  Our training settings are also the same with DARTS, we stack  $6$ normal cells and $2$ reduction cells to build a network, and the reduction cells are inserted in a network of $1/3$ and $2/3$, respectively. 	 A cell includes $7$ nodes with $4$ intermediate nodes, every node represents a feature map, and the number of edges in a cell is $14$. 

 We perform an architecture search for $50$ epochs with a batch size of $64$, and the architecture search is divided into $8$ stages. Each  stage includes $2$ epochs. We increase one candidate operation in the search space at the beginning of each stage and train the supernet until the next stage.  After the final stage, we train the supernet until the final epoch. During the training, we leverage SGD \cite{QIAN1999145} as an optimizer to optimize model weights $W$, and its initial learning rate is $3 \times 10^{-4}$, momentum is $(0.5, 0.999)$,   weight decay is $10^{-3}$.  At the same time, we leverage Adam \cite{kingma2017adam} to optimize architecture weights, and its  initial learning rate is $3 \times 10^{-4}$, momentum is $(0.5, 0.999)$,   weight decay is $10^{-3}$.  The search phase spends 0.4 GPU day on a single NVIDIA GTX 2080Ti. We ran $5$ times independent search experiments with 5 different seeds, and the best cells were found are illustrated in Figure \ref{fig_4}.

\subsection{Architecture Evaluation on CIFAR-10}
\label{ae_10}
\begin{table}[]
	\centering
	\scriptsize
	\caption{Comparison with state-of-the-art image classifiers on CIFAR-10 (lower error rate is better).}
	\label{tb1}
	\begin{threeparttable}
		\begin{tabular}{@{}ccccc@{}}
			\toprule
			\textbf{Architecture} &
			\begin{tabular}[c]{@{}c@{}} \textbf{Test Err.}\\ (\textbf{\%})\end{tabular} &
			\begin{tabular}[c]{@{}c@{}}\textbf{Params}\\ \textbf{(M)}\end{tabular} &
			\multicolumn{1}{l}{\begin{tabular}[c]{@{}l@{}}\textbf{Search Cost}\\ \textbf{(GPU-days)}\end{tabular}} &
			\begin{tabular}[c]{@{}c@{}}\textbf{Search} \\ \textbf{Method}\end{tabular} \\ \midrule
			DenseNet-BC \cite{8099726}    & 3.46  & 25.6 & -    & manual    \\ \midrule
			NASNet-A \cite{2017arXiv170707012Z} & 2.65     & 3.3  & 1800 & RL        \\
			AmoebaNet-A \cite{2018arXiv180201548R}    & 3.34$\pm$0.06 & 3.2  & 3150 & evolution \\
			AmoebaNet-B \cite{2018arXiv180201548R}    & 2.55$\pm$0.05 & 2.8  & 3150 & evolution \\
			PNAS  \cite{2017arXiv171200559L}        & 3.41$\pm$0.09 & 3.2  & 225  & SMBO      \\
			ENAS \cite{enas}          & 2.89          & 4.6  & 0.5  & RL        \\ \midrule
			DARTS ($1^{\text{st}}$ order)  \cite{2018arXiv180609055L} & 3.00$\pm$0.14 & 3.3  & 0.4  & gradient  \\
			DARTS ($2^{\text{nd}}$ order) \cite{2018arXiv180609055L} & 2.76$\pm$0.09 & 3.3  & 1    & gradient  \\
			SNAS (mild)  \cite{2018arXiv181209926X}      & 2.98          & 2.9  & 1.5  & gradient  \\
			ProxylessNAS  \cite{cai2018proxylessnas}     & 2.08          & -    & 4    & gradient  \\
			P-DARTS  \cite{Chen_2019_ICCV}       & 2.5           & 3.4  & 0.3  & gradient  \\
			PC-DARTS  \cite{Xu2020PC-DARTS:}     & 2.57$\pm$0.07 & 3.6  & 0.1  & gradient  \\
			SDARTS-RS \cite{pmlr-v119-chen20f}     & 2.67$\pm$0.03 & 3.4  & 0.4  & gradient  \\
			GDAS  \cite{8953848}     & 2.93          & 3.4  & 0.3  & gradient  \\
			R-DARTS (L2) \cite{Zela2020Understanding}       & 2.95$\pm$0.21 & -    & 1.6  & gradient  \\
			SGAS (Cri 1. avg)  \cite{li2019sgas}    & 2.66$\pm$0.24 & 3.7  & 0.25 & gradient  \\
			DARTS-PT \cite{wang2021rethinking}        & 2.61$\pm$0.08 & 3.0  & 0.8  & gradient  \\ \midrule
			OPP-DARTS     & 2.63$\pm$0.27 \tnote{1} & 3.8  & 0.4 \tnote{2}  & gradient  \\ \bottomrule 	
		\end{tabular}
		
		\begin{tablenotes}
			\footnotesize
			\item[1] We ran OPP-DARTS $5$ times with different search seeds to search cells, and evaluated  the best cells $10$ times with different evaluation seeds to get average test error and variance of test error.
			\item[2] Recorded on a single GTX 2080Ti. 
		\end{tablenotes}
		
	\end{threeparttable}
\end{table}

The cells found by OPP-DARTS have shown in Figure \ref{fig_4}, and then we will stack these cells to evaluate their performance on CIFAR-10.  We build a large network with $20$ cells and $36$ initial channels. After building the network, we train the stacked network until $600$ epochs with a batch size of $96$. Furthermore, other training settings are the same with DARTS, such as cutout with length $16$,  auxiliary towers with weight $0.4$, and path dropout with a probability of $0.3$.  

We compare our result on CIFAR-10 with other methods in Table \ref{tb1}, such as Darts- \cite{chu2021darts}, SGAS \cite{li2019sgas}. Table \ref{tb1} shows that our method (OPP-DARTS)  improves standard DARTS' test error from $3.00\%$ to $2.63\%$, with identical search cost ($0.4$ GPU days) on a single NVIDIA GTX 2080Ti.  Our method surpasses them a lot compared to other indicator-based variants, such as SDARTS-RS and R-DARTS (L2).  The result further demonstrates that the  indicator-based method can alleviate  skip connections aggregation but prevent DARTS from exploring better architectures.

\subsection{Architecture Evaluation on ImageNet}
\label{ae_in}

To verify the transferability of our method, we evaluate architecture on ImageNet by using the best cells searched on CIFAR-10. ILSVRC 2012 \cite{ILSVRC15} is a famous ImageNet dataset, and we leverage it to test our architecture searched on CIFAR-10. 

The network configuration is also the same as DARTS, i.e., the network used to evaluate is stacked by $14$ cells, and the number of its initial channels is $48$. The stacked network is trained from scratch for 250 epochs, and its batch size is $256$ during the training phase of the network. An SGD optimizer optimizes the network's parameters;  the optimizer's initial learning rate is $0.2$, weight decay is $3 \times 10^{-5}$, momentum is $0.9$.


\begin{table}[]
	\centering
	\scriptsize
	\setlength\tabcolsep{3pt}
	\caption{Comparison with state-of-the-art classifiers on ImageNet.}
	\label{tb2}
	\begin{threeparttable}
		\begin{tabular}{@{}ccccccc@{}}
			\toprule
			\multirow{2}{*}{\textbf{Architecture}} &
			\multicolumn{2}{c}{\textbf{Test Err.(\%)}} &
			\multirow{2}{*}{\begin{tabular}[c]{@{}c@{}}\textbf{Params}\\ \textbf{(M)}\end{tabular}} &
			\multirow{2}{*}{\begin{tabular}[c]{@{}c@{}}$\times +$\\ \textbf{(M)}\end{tabular}} &
			\multirow{2}{*}{\begin{tabular}[c]{@{}c@{}}\textbf{Search Cost}\\ \textbf{(GPU-days)}\end{tabular}} &
			\multirow{2}{*}{\begin{tabular}[c]{@{}c@{}}\textbf{Search}\\ \textbf{Method}\end{tabular}} \\ \cline{2-3}
			& \textbf{top-1} & \textbf{top-5} &  &  &  &  \\ \midrule
			Inception-v1 \cite{szegedy2014going}         & 30.2          & 10.1              & 6.6     & 1448 & -    & manual    \\
			MobileNet \cite{howard2017mobilenets}   & 29.4                      & 10.5                      & 4.2     & 569  & -    & manual    \\
			ShuffleNet 2x (v1) \cite{8578814}   & 26.4                      & 10.2                      & $\sim$5 & 524  & -    & manual    \\
			ShuffleNet 2x (v2)  \cite{10.1007/978-3-030-01264-9_8}  & 25.1                      & -                         & $\sim$5 & 591  & -    & manual    \\ \midrule
			NASNet-A  \cite{2017arXiv170707012Z}            & 26                        & 8.4                       & 5.3     & 564  & 1800 & RL        \\
			NASNet-B   \cite{2017arXiv170707012Z}           & 27.2                      & 8.7                       & 5.3     & 488  & 1800 & RL        \\
			NASNet-C   \cite{2017arXiv170707012Z}           & 27.5                      & 9                         & 4.9     & 558  & 1800 & RL        \\
			AmoebaNet-A  \cite{2018arXiv180201548R}         & 25.5                      & 8                         & 5.1     & 555  & 3150 & evolution \\
			AmoebaNet-B    \cite{2018arXiv180201548R}       & 26                        & 8.5                       & 5.3     & 555  & 3150 & evolution \\
			AmoebaNet-C    \cite{2018arXiv180201548R}       & 24.3                      & 7.6                       & 6.4     & 570  & 3150 & evolution \\
			FairNAS-A  \cite{chu2020fairnas}           & 24.7                      & 7.6                       & 4.6     & 388  & 12   & evolution \\
			PNAS   \cite{2017arXiv171200559L}               & 25.8                      & 8.1                       & 5.1     & 588  & 225  & SMBO      \\
			MnasNet-92 \cite{8954198}           & 25.2                      & 8                         & 4.4     & 388  & -    & RL        \\ \midrule
			DARTS($2^{\text{nd}}$ order) \cite{2018arXiv180609055L} & 26.7                      & 8.7                       & 4.7     & 574  & 4.0  & gradient  \\
			SNAS (mild)  \cite{2018arXiv181209926X}          & 27.3                      & 9.2                       & 4.3     & 522  & 1.5  & gradient  \\
			ProxylessNAS \cite{cai2018proxylessnas}         & 24.9                      & 7.5                       & 7.1     & 465  & 8.3  & gradient  \\
			P-DARTS \cite{Chen_2019_ICCV}              & 24.4                      & 7.4                       & 4.9     & 557  & 0.3  & gradient  \\
			PC-DARTS  \cite{Xu2020PC-DARTS:}             & 25.1                      & 7.8                       & 5.3     & 586  & 0.1  & gradient  \\
			SGAS (Cri.1 avg.)  \cite{li2019sgas}   & 24.41                     & 7.29                      & 5.3     & 579  & 0.25 & gradient  \\
			GDAS         \cite{8953848}         & 26.0                      & 8.5                       & 5.3     & 581  & 0.21 & gradient  \\ \midrule
			OPP-DARTS   &    25.61     &      8.15        & 5.3     &  610    & 0.4  & gradient  \\ \bottomrule
		\end{tabular}
	\end{threeparttable}
\end{table}

Table \ref{tb2} shows our evaluation result, and we compare our result with SOTA manual architectures and models obtained through other search methods.  The architecture found by OPP-DARTS on CIFAR-10  is superior to the architecture found by standard DARTS in the image classification task, and it  means that the transferability of our method is better  than standard DARTS.

\subsection{ Robustness of OPP-DARTS}
\label{rbt}

Because of skip connections aggregation, DARTS will face  a performance crash when DARTS searches  architectures on three simple search space \cite{Zela2020Understanding}, i.e., \textbf{S2}, \textbf{S3}, \textbf{S4}. \textbf{S2} is consist of two operations per edge: $(skip\_connect, 3 \times 3\_sep\_conv)$. \textbf{S3}  is consist of three operations per edge: $(skip\_connect, 3 \times 3\_sep\_conv, zero)$. \textbf{S4}  is consist of two operations per edge: $(noise, 3 \times 3\_sep\_conv)$.   \cite{Zela2020Understanding} discovers that when DARTS is used to search architectures on \textbf{S2}, \textbf{S3}, it finds suboptimal architectures, and the performance of architectures is inferior. Even though on \textbf{S4}, DARTS selects excessive harmful $noise$ operations.

To verify the robustness of our methods, we search architectures on these search spaces by OPP-DARTS. Table \ref{tb3} shows the results of the OPP-DARTS search in the \textbf{S2}-\textbf{S4}. When searching on \textbf{S2}, the performance of  OPP-DARTS (a test error of $2.84\%$) is far further than   DARTS.  At the same time,  we use OPP-DARTS to search architectures on \textbf{S3} and \textbf{S4}. Table \ref{tb3} also shows that the performance of OPP-DARTS on \textbf{S3} and \textbf{S4} is better than DARTS. Based on these above experiments, OPP-DARTS is more robust than DARTS.

\begin{table}[]
	\centering
	\caption{OPP-DARTS on \textbf{S2}-\textbf{S4} (test error ($\%$))}
	\label{tb3}
	\begin{tabular}{c|c|c}
		\hline
		\textbf{Space} & \textbf{DARTS} & \textbf{OPP-DARTS} \\ \hline
		\textbf{S2}    & 5.94  & 2.93      \\ \hline
		\textbf{S3}    & 3.04  & 2.90      \\ \hline
		\textbf{S4}    & 4.85  & 3.21      \\ \hline
	\end{tabular}
\end{table}

\section{Conclusions}

In this paper, we propose operation-level progressive differentiable architecture search to alleviate skip connections aggregation. The core idea is that we split the search space into multiple stages and increase operations into search space at the beginning of each stage. In addition, we select the operation that makes the training loss of the supernet largest from candidate operations set to keep operations competing fairly.

\section{Acknowledgements}

 This work is supported in part by Excellent Talents Program of Institute of Information Engineering, CAS, Special Research Assistant Project of CAS (No. E0YY231114), Beijing Outstanding Young Scientist Program (No. BJJWZYJH012019100020098) and National Natural Science Foundation of China (No. 62076234, No. 62106257). The work is also supported by Intelligent Social Governance Platform, Major Innovation \& Planning Interdisciplinary Platform for the “Double-First Class” Initiative, Renmin University of China. We also wish to acknowledge the support provided and contribution made by Public Policy and Decision-making Research Lab of Renmin University of China and China Unicom Innovation Ecological Cooperation Plan.

\bibliographystyle{IEEEtran}
\bibliography{ref}{}

\begin{thebibliography}{10}
\providecommand{\url}[1]{#1}
\csname url@samestyle\endcsname
\providecommand{\newblock}{\relax}
\providecommand{\bibinfo}[2]{#2}
\providecommand{\BIBentrySTDinterwordspacing}{\spaceskip=0pt\relax}
\providecommand{\BIBentryALTinterwordstretchfactor}{4}
\providecommand{\BIBentryALTinterwordspacing}{\spaceskip=\fontdimen2\font plus
\BIBentryALTinterwordstretchfactor\fontdimen3\font minus
  \fontdimen4\font\relax}
\providecommand{\BIBforeignlanguage}[2]{{%
\expandafter\ifx\csname l@#1\endcsname\relax
\typeout{** WARNING: IEEEtran.bst: No hyphenation pattern has been}%
\typeout{** loaded for the language `#1'. Using the pattern for}%
\typeout{** the default language instead.}%
\else
\language=\csname l@#1\endcsname
\fi
#2}}
\providecommand{\BIBdecl}{\relax}
\BIBdecl

\bibitem{DBLP:conf/aaai/LiLW20}
\BIBentryALTinterwordspacing
J.~Li, Y.~Liu, and W.~Wang, ``Automated spectral kernel learning,'' in
  \emph{The Thirty-Fourth {AAAI} Conference on Artificial Intelligence, {AAAI}
  2020, The Thirty-Second Innovative Applications of Artificial Intelligence
  Conference, {IAAI} 2020, The Tenth {AAAI} Symposium on Educational Advances
  in Artificial Intelligence, {EAAI} 2020, New York, NY, USA, February 7-12,
  2020}.\hskip 1em plus 0.5em minus 0.4em\relax {AAAI} Press, 2020, pp.
  4618--4625. [Online]. Available:
  \url{https://aaai.org/ojs/index.php/AAAI/article/view/5892}
\BIBentrySTDinterwordspacing

\bibitem{DBLP:conf/ijcai/LiLY019}
\BIBentryALTinterwordspacing
J.~Li, Y.~Liu, R.~Yin, and W.~Wang, ``Multi-class learning using unlabeled
  samples: Theory and algorithm,'' in \emph{Proceedings of the Twenty-Eighth
  International Joint Conference on Artificial Intelligence, {IJCAI} 2019,
  Macao, China, August 10-16, 2019}, S.~Kraus, Ed.\hskip 1em plus 0.5em minus
  0.4em\relax ijcai.org, 2019, pp. 2880--2886. [Online]. Available:
  \url{https://doi.org/10.24963/ijcai.2019/399}
\BIBentrySTDinterwordspacing

\bibitem{DBLP:conf/icml/0001WM21}
\BIBentryALTinterwordspacing
R.~Yin, Y.~Liu, W.~Wang, and D.~Meng, ``Distributed nystr{\"{o}}m kernel
  learning with communications,'' in \emph{Proceedings of the 38th
  International Conference on Machine Learning, {ICML} 2021, 18-24 July 2021,
  Virtual Event}, ser. Proceedings of Machine Learning Research, M.~Meila and
  T.~Zhang, Eds., vol. 139.\hskip 1em plus 0.5em minus 0.4em\relax {PMLR},
  2021, pp. 12\,019--12\,028. [Online]. Available:
  \url{http://proceedings.mlr.press/v139/yin21a.html}
\BIBentrySTDinterwordspacing

\bibitem{726791}
Y.~Lecun, L.~Bottou, Y.~Bengio, and P.~Haffner, ``Gradient-based learning
  applied to document recognition,'' \emph{Proceedings of the IEEE}, vol.~86,
  no.~11, pp. 2278--2324, 1998.

\bibitem{NIPS2012_c399862d}
A.~Krizhevsky, I.~Sutskever, and G.~E. Hinton, ``Imagenet classification with
  deep convolutional neural networks,'' in \emph{Advances in Neural Information
  Processing Systems}, F.~Pereira, C.~J.~C. Burges, L.~Bottou, and K.~Q.
  Weinberger, Eds., vol.~25.\hskip 1em plus 0.5em minus 0.4em\relax Curran
  Associates, Inc., 2012.

\bibitem{simonyan2015deep}
K.~Simonyan and A.~Zisserman, ``Very deep convolutional networks for
  large-scale image recognition,'' 2015.

\bibitem{45826}
\BIBentryALTinterwordspacing
B.~Zoph and Q.~V. Le, ``Neural architecture search with reinforcement
  learning,'' 2017. [Online]. Available: \url{https://arxiv.org/abs/1611.01578}
\BIBentrySTDinterwordspacing

\bibitem{real2019regularized}
E.~Real, A.~Aggarwal, Y.~Huang, and Q.~V. Le, ``Regularized evolution for image
  classifier architecture search,'' 2019.

\bibitem{2017arXiv170707012Z}
B.~{Zoph}, V.~{Vasudevan}, J.~{Shlens}, and Q.~V. {Le}, ``{Learning
  Transferable Architectures for Scalable Image Recognition},'' \emph{arXiv
  e-prints}, p. arXiv:1707.07012, Jul. 2017.

\bibitem{Zela2020Understanding}
\BIBentryALTinterwordspacing
A.~Zela, T.~Elsken, T.~Saikia, Y.~Marrakchi, T.~Brox, and F.~Hutter,
  ``Understanding and robustifying differentiable architecture search,'' in
  \emph{International Conference on Learning Representations}, 2020. [Online].
  Available: \url{https://openreview.net/forum?id=H1gDNyrKDS}
\BIBentrySTDinterwordspacing

\bibitem{pmlr-v119-chen20f}
\BIBentryALTinterwordspacing
X.~Chen and C.-J. Hsieh, ``Stabilizing differentiable architecture search via
  perturbation-based regularization,'' in \emph{Proceedings of the 37th
  International Conference on Machine Learning}, ser. Proceedings of Machine
  Learning Research, H.~D. III and A.~Singh, Eds., vol. 119.\hskip 1em plus
  0.5em minus 0.4em\relax PMLR, 13--18 Jul 2020, pp. 1554--1565. [Online].
  Available: \url{http://proceedings.mlr.press/v119/chen20f.html}
\BIBentrySTDinterwordspacing

\bibitem{Chen_2019_ICCV}
X.~Chen, L.~Xie, J.~Wu, and Q.~Tian, ``Progressive differentiable architecture
  search: Bridging the depth gap between search and evaluation,'' in
  \emph{Proceedings of the IEEE/CVF International Conference on Computer Vision
  (ICCV)}, October 2019.

\bibitem{chu2021darts}
\BIBentryALTinterwordspacing
X.~Chu, X.~Wang, B.~Zhang, S.~Lu, X.~Wei, and J.~Yan, ``{\{}DARTS{\}}-:
  Robustly stepping out of performance collapse without indicators,'' in
  \emph{International Conference on Learning Representations}, 2021. [Online].
  Available: \url{https://openreview.net/forum?id=KLH36ELmwIB}
\BIBentrySTDinterwordspacing

\bibitem{2017arXiv171200559L}
C.~{Liu}, B.~{Zoph}, M.~{Neumann}, J.~{Shlens}, W.~{Hua}, L.-J. {Li},
  L.~{Fei-Fei}, A.~{Yuille}, J.~{Huang}, and K.~{Murphy}, ``{Progressive Neural
  Architecture Search},'' \emph{arXiv e-prints}, p. arXiv:1712.00559, Dec.
  2017.

\bibitem{2018arXiv180201548R}
E.~{Real}, A.~{Aggarwal}, Y.~{Huang}, and Q.~V. {Le}, ``{Regularized Evolution
  for Image Classifier Architecture Search},'' \emph{arXiv e-prints}, p.
  arXiv:1802.01548, Feb. 2018.

\bibitem{2018arXiv180609055L}
H.~{Liu}, K.~{Simonyan}, and Y.~{Yang}, ``{DARTS: Differentiable Architecture
  Search},'' \emph{arXiv e-prints}, p. arXiv:1806.09055, Jun. 2018.

\bibitem{wang2021rethinking}
\BIBentryALTinterwordspacing
R.~Wang, M.~Cheng, X.~Chen, X.~Tang, and C.-J. Hsieh, ``Rethinking architecture
  selection in differentiable {NAS},'' in \emph{International Conference on
  Learning Representations}, 2021. [Online]. Available:
  \url{https://openreview.net/forum?id=PKubaeJkw3}
\BIBentrySTDinterwordspacing

\bibitem{krizhevsky2009learning}
A.~Krizhevsky, G.~Hinton \emph{et~al.}, ``Learning multiple layers of features
  from tiny images,'' 2009.

\bibitem{QIAN1999145}
N.~Qian, ``On the momentum term in gradient descent learning algorithms,''
  \emph{Neural Networks}, vol.~12, no.~1, pp. 145--151, 1999.

\bibitem{kingma2017adam}
D.~P. Kingma and J.~Ba, ``Adam: A method for stochastic optimization,'' 2017.

\bibitem{8099726}
G.~Huang, Z.~Liu, L.~Van Der~Maaten, and K.~Q. Weinberger, ``Densely connected
  convolutional networks,'' in \emph{2017 IEEE Conference on Computer Vision
  and Pattern Recognition (CVPR)}, 2017, pp. 2261--2269.

\bibitem{enas}
H.~Pham, M.~Y. Guan, B.~Zoph, Q.~V. Le, and J.~Dean, ``Efficient neural
  architecture search via parameter sharing,'' in \emph{ICML}, 2018.

\bibitem{2018arXiv181209926X}
S.~{Xie}, H.~{Zheng}, C.~{Liu}, and L.~{Lin}, ``{SNAS: Stochastic Neural
  Architecture Search},'' \emph{arXiv e-prints}, p. arXiv:1812.09926, Dec.
  2018.

\bibitem{cai2018proxylessnas}
\BIBentryALTinterwordspacing
H.~Cai, L.~Zhu, and S.~Han, ``Proxyless{NAS}: Direct neural architecture search
  on target task and hardware,'' in \emph{International Conference on Learning
  Representations}, 2019. [Online]. Available:
  \url{https://openreview.net/forum?id=HylVB3AqYm}
\BIBentrySTDinterwordspacing

\bibitem{Xu2020PC-DARTS:}
\BIBentryALTinterwordspacing
Y.~Xu, L.~Xie, X.~Zhang, X.~Chen, G.-J. Qi, Q.~Tian, and H.~Xiong, ``Pc-darts:
  Partial channel connections for memory-efficient architecture search,'' in
  \emph{International Conference on Learning Representations}, 2020. [Online].
  Available: \url{https://openreview.net/forum?id=BJlS634tPr}
\BIBentrySTDinterwordspacing

\bibitem{8953848}
X.~Dong and Y.~Yang, ``Searching for a robust neural architecture in four gpu
  hours,'' in \emph{2019 IEEE/CVF Conference on Computer Vision and Pattern
  Recognition (CVPR)}, 2019, pp. 1761--1770.

\bibitem{li2019sgas}
G.~Li, G.~Qian, I.~C. Delgadillo, M.~M{\"u}ller, A.~Thabet, and B.~Ghanem,
  ``Sgas: Sequential greedy architecture search,'' in \emph{Proceedings of IEEE
  Conference on Computer Vision and Pattern Recognition (CVPR)}, 2020.

\bibitem{ILSVRC15}
O.~Russakovsky, J.~Deng, H.~Su, J.~Krause, S.~Satheesh, S.~Ma, Z.~Huang,
  A.~Karpathy, A.~Khosla, M.~Bernstein, A.~C. Berg, and L.~Fei-Fei, ``{ImageNet
  Large Scale Visual Recognition Challenge},'' \emph{International Journal of
  Computer Vision (IJCV)}, vol. 115, no.~3, pp. 211--252, 2015.

\bibitem{szegedy2014going}
C.~Szegedy, W.~Liu, Y.~Jia, P.~Sermanet, S.~Reed, D.~Anguelov, D.~Erhan,
  V.~Vanhoucke, and A.~Rabinovich, ``Going deeper with convolutions,'' 2014.

\bibitem{howard2017mobilenets}
A.~G. Howard, M.~Zhu, B.~Chen, D.~Kalenichenko, W.~Wang, T.~Weyand,
  M.~Andreetto, and H.~Adam, ``Mobilenets: Efficient convolutional neural
  networks for mobile vision applications,'' 2017.

\bibitem{8578814}
X.~Zhang, X.~Zhou, M.~Lin, and J.~Sun, ``Shufflenet: An extremely efficient
  convolutional neural network for mobile devices,'' in \emph{2018 IEEE/CVF
  Conference on Computer Vision and Pattern Recognition}, 2018, pp. 6848--6856.

\bibitem{10.1007/978-3-030-01264-9_8}
N.~Ma, X.~Zhang, H.-T. Zheng, and J.~Sun, ``Shufflenet v2: Practical guidelines
  for efficient cnn architecture design,'' in \emph{Computer Vision -- ECCV
  2018}, V.~Ferrari, M.~Hebert, C.~Sminchisescu, and Y.~Weiss, Eds.\hskip 1em
  plus 0.5em minus 0.4em\relax Cham: Springer International Publishing, 2018,
  pp. 122--138.

\bibitem{chu2020fairnas}
X.~Chu, B.~Zhang, R.~Xu, and J.~Li, ``Fairnas: Rethinking evaluation fairness
  of weight sharing neural architecture search,'' 2020.

\bibitem{8954198}
\BIBentryALTinterwordspacing
M.~Tan, B.~Chen, R.~Pang, V.~Vasudevan, M.~Sandler, A.~Howard, and Q.~V. Le,
  ``Mnasnet: Platform-aware neural architecture search for mobile,'' in
  \emph{2019 IEEE/CVF Conference on Computer Vision and Pattern Recognition
  (CVPR)}.\hskip 1em plus 0.5em minus 0.4em\relax Los Alamitos, CA, USA: IEEE
  Computer Society, jun 2019, pp. 2815--2823. [Online]. Available:
  \url{https://doi.ieeecomputersociety.org/10.1109/CVPR.2019.00293}
\BIBentrySTDinterwordspacing

\end{thebibliography}

\end{document}